\documentclass[letterpaper, 10 pt, conference]{ieeeconf}  

\IEEEoverridecommandlockouts                              

\usepackage{graphicx} 
\usepackage{amsmath} 
\usepackage{amssymb}  
\usepackage{bm}
\usepackage{lipsum}
\usepackage{mathtools}
\usepackage{scalerel}

\usepackage{url}

\usepackage{array,multirow}

\let\labelindent\relax
\usepackage{enumitem}

\usepackage{placeins} 
\usepackage{flafter}  

\usepackage{balance}
\usepackage{xcolor}
\usepackage{setspace}

\usepackage[linesnumbered,ruled,vlined]{algorithm2e}

\SetCommentSty{mycommfont}
\SetKwInput{KwInput}{Input}                
\SetKwInput{KwOutput}{Output}  
\SetKwInput{KwParam}{Parameter}

\usepackage{lipsum}

\usepackage[hidelinks]{hyperref}

\title{\LARGE \bf
PCR-99: A Practical Method for Point Cloud Registration \\ with 99 Percent Outliers 
}

\author{Seong Hun Lee$^{1}$, Javier Civera$^{2}$ and Patrick Vandewalle$^{3}$
\thanks{*This work was supported by the Spanish
Government (projects PID2021-127685NB-I00 and TED2021-131150BI00) and the Arag\'on Government
(project DGA-T45 23R).}
\thanks{$^{1,2}$ Seong Hun Lee and Javier Civera are with I3A, University of Zaragoza, 50018 Zaragoza, Spain.
{\tt\small seonghunlee@unizar.es}, {\tt\small jcivera@unizar.es}}%
\thanks{$^{3}$ Patrick Vandewalle is with EAVISE, Department of Electrical Engineering, KU Leuven, 2860 Sint-Katelijne-Waver, Belgium.
        {\tt\small patrick.vandewalle@kuleuven.be}}%
}

\begin{document}

\maketitle
\thispagestyle{empty}
\pagestyle{empty}

\begin{abstract}
We propose a robust method for point cloud registration that can handle both unknown scales and extreme outlier ratios.
Our method, dubbed PCR-99, uses a deterministic 3-point sampling approach with two novel mechanisms that significantly boost the speed: (1) an improved ordering of the samples based on pairwise scale consistency,  prioritizing the point correspondences that are more likely to be inliers, and (2) an efficient outlier rejection scheme based on triplet scale consistency, prescreening bad samples and reducing the number of hypotheses to be tested.
Our evaluation shows that, up to 98\% outlier ratio, the proposed method achieves comparable performance to the state of the art.
At 99\% outlier ratio, however, it outperforms the state of the art for both known-scale and unknown-scale problems.
Especially for the latter, we observe a clear superiority in terms of robustness and speed.
\end{abstract}

\section{INTRODUCTION}
\label{sec:intro}

\begin{figure}[t]
 \centering
 \includegraphics[width=0.4\textwidth]{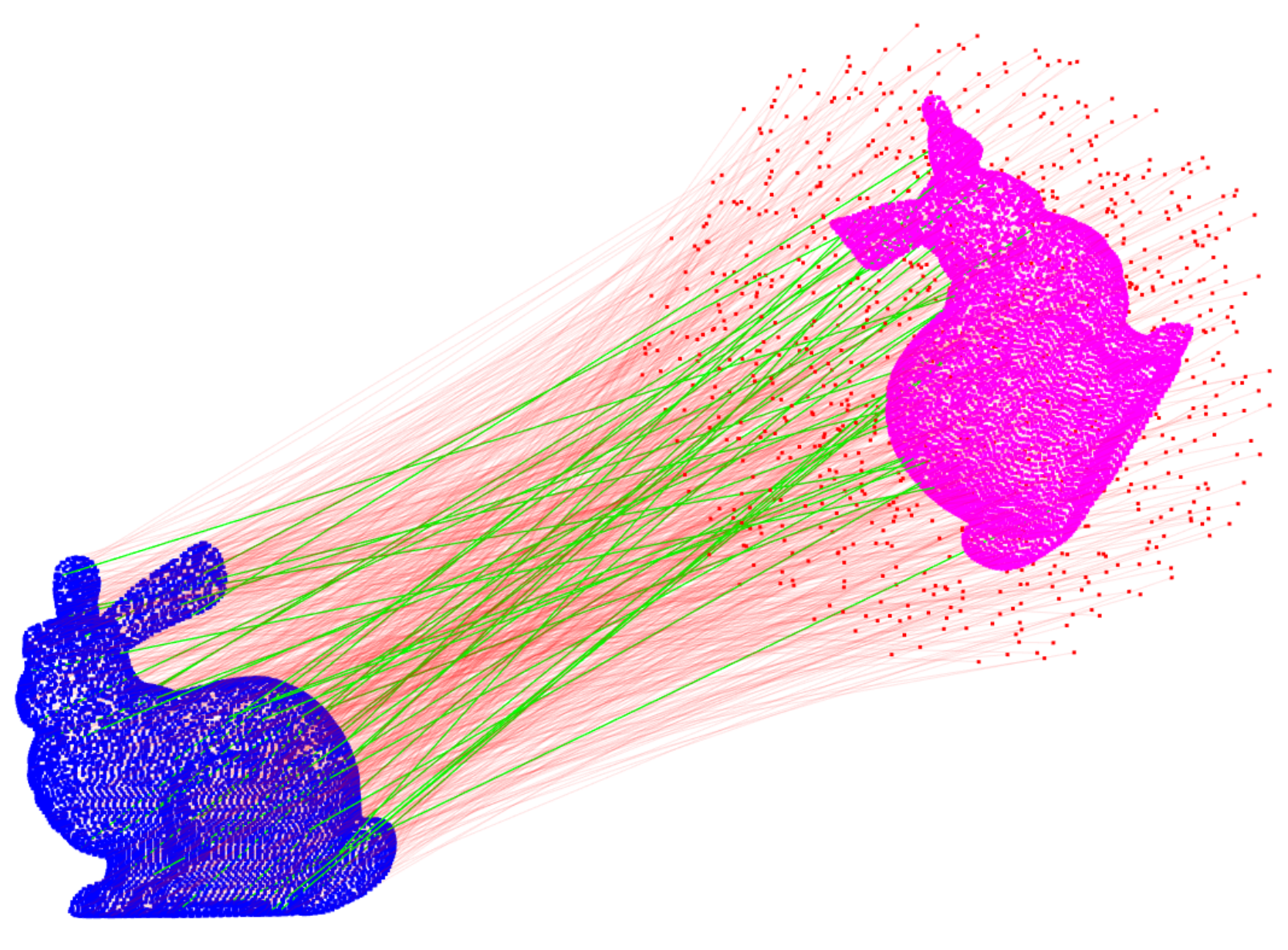}
\caption{The Bunny dataset \cite{bunny} used in our evaluation.
The inlier correspondences are shown as the thick green lines and the outliers as the thin red lines.
Here, the outlier ratio is 96\%. 
Image credit: \cite{sra2}.}
\label{fig:bunny}
\end{figure}

We consider the problem of correspondence-based point cloud registration with or without the knowledge of the relative scale, \textit{i.e.}, aligning two corresponding sets of noisy and outlier-contaminated 3D points to obtain the most accurate relative rotation, translation (and scale) between them.
This problem is relevant in many applications, such as 3D scene reconstruction \cite{henry_2012_rgbd, choi_2015_cvpr}, object recognition and localization \cite{drost_2010_cvpr, marion_2018_icra}, and simultaneous localization and mapping \cite{zhang_2014_rss}.

Unlike the commonly used Iterative Closest Point (ICP) \cite{besl_1992_tpami}, correspondence-based methods have the advantage that they do not rely on the initial guess for the unknown transformation between the two point sets.
Instead, they first match 3D keypoints using feature descriptors (\textit{e.g.}, FPFH \cite{rusu_icra_2009}), and then infer the transformation from the putative correspondences.
In practice, however, these correspondences may occasionally contain a large proportion of outliers (\textit{e.g.} more than 95\%) \cite{bustos_2018_tpami}.
It is therefore important that the estimation method can handle such extreme outlier ratios.

In this work, we propose a novel point cloud registration method that is robust to an extremely large amount of outliers (up to 99\%).
Our main contribution is two-fold:
\begin{enumerate}
    \item We develop a deterministic approach for choosing 3-point samples in such a way that the inlier correspondences are more likely to be chosen earlier than the outliers.
    The key idea is to rank each point based on \textit{pairwise scale consistency} and prioritize samples with a smaller total ranking number.
    \item We propose an efficient outlier rejection scheme that enables us to prescreen bad samples without having to compute the associated hypotheses, \textit{i.e.}, rigid body transformation for known-scale problems and similarity transformation for unknown-scale problems.
    The key idea is to evaluate the \textit{triplet scale consistency} of a given sample using precomputed data.
\end{enumerate}
We conduct a thorough evaluation and demonstrate that our method leads to state-of-the-art results in point cloud registration up to 99\% outlier ratio.
Our code is publicly available at 
{\color{magenta}\url{https://github.com/sunghoon031/PCR-99}}.

The remainder of this paper is organized as follows:
First, related work is discussed in Section \ref{sec:related}.
We formulate the problem in Section \ref{sec:problem} and provide an overview of our method in Section \ref{sec:overview}.
Section \ref{sec:unknown_scale} and \ref{sec:known_scale} describe our method for unknown-scale and known-scale problems, respectively.
We present the evaluation results in Section \ref{sec:results} and draw conclusions in Section \ref{sec:conclusion}.

\section{RELATED WORK}
\label{sec:related}
In this section, we briefly review some of the existing robust correspondence-based methods. 

\textit{Consensus Maximization Methods}:
Two popular methods that maximize the consensus set are RANSAC and Branch-and-Bound (BnB).
RANSAC \cite{ransac} and its variants (\textit{e.g.}, \cite{chum_2003_pr, nister_2003_iccv}) maximize the consensus by taking minimal samples randomly and iteratively.
BnB \cite{horst_2013_global, bustos_2016_tpami} finds the globally optimal solution by searching directly in the parameter space.
Both RANSAC and BnB run in exponential time in the worst case \cite{teaser}.
Recently, however, several RANSAC-based methods have been shown to achieve impressive robustness and speed, even at 99\% outlier ratio (\textit{e.g.}, \cite{ransic, icos, daniel, vodrac}).
In \cite{bustos_2018_tpami}, a technique called guaranteed outlier removal (GORE) is proposed.
This method finds the globally optimal solution, but it scales poorly, as it uses BnB as a possible subroutine.
In \cite{qgore}, a quadratic-time GORE algorithm is proposed, which is shown to be more efficient.
Our method is most similar to RANSAC-based consensus maximization methods, in that we also try to find the minimal sample of inliers iteratively.
The main difference is that our sampling is done deterministically.

\textit{M-estimation Methods}:
M-estimation minimizes robust cost functions, down-weighting the effect of outliers iteratively \cite{g2o, mactavish2015all}.
This method is sensitive to the initial guess and can easily get stuck in local minima.
To reduce this sensitivity and increase the robustness to outliers, graduated non-convexity (GNC) can be used \cite{gnc, fgr}.
However, this type of local iterative method is not capable of handling extreme outlier ratios (\textit{e.g.}, $>90\%$).

\textit{Graph-Based Methods}:
Some methods use graph theory to achieve robust point cloud registration. 
For example, in \cite{teaser, yang_2019_rss, robin}, graph theory is used to identify inliers based on invariance constraints. 
Readers interested in more recent graph-based methods are referred to \cite{pointdsc,sc2pcr,mac,pointmc}.

\textit{Single Rotation Averaging}: Single rotation averaging (\textit{e.g.}, \cite{sra}) has been used to maximize the consensus set \cite{vocra} or to find the rotation separately \cite{sra2} in point cloud registration.

\section{PROBLEM DEFINITION}
\label{sec:problem}
Let $\{\mathbf{a}_i\}_{i=1}^{n}$ and $\{\mathbf{b}_i\}_{i=1}^{n}$ be two sets of the same number of 3D points.
Let $\{\mathbf{a}_i\}_{i\in\mathcal{I}}$ and $\{\mathbf{b}_i\}_{i\in\mathcal{I}}$ be two subsets related by a similarity transformation with some noise $\{\mathbf{e}_i\}_{i\in\mathcal{I}}$, \textit{i.e.},
\begin{equation}
\label{eq:sim_from_3}
    s\mathbf{R}\mathbf{a}_i+\mathbf{t} +\mathbf{e}_i = \mathbf{b}_i \quad \text{for} \quad i \in \mathcal{I},
\end{equation}
where $\mathcal{I}$ is the set of inlier point correspondences.
Their complements, $\{\mathbf{a}_i\}_{i\notin\mathcal{I}}$ and $\{\mathbf{b}_i\}_{i\notin\mathcal{I}}$, are not related to each other and they are considered to be the outliers.
Given that $\mathcal{I}$ is unknown, estimating $(s, \mathbf{R}, \mathbf{t})$ from $\{\mathbf{a}_i\}_{i=1}^{n}$ and $\{\mathbf{b}_i\}_{i=1}^{n}$ is called the \textit{unknown-scale} point cloud registration problem.
If $s$ is known a priori, it is called the \textit{known-scale} problem.
In this work, we consider both problems.

\section{OVERVIEW OF PCR-99}
\label{sec:overview}
For both known-scale and unknown-scale cases, our method consists of the following steps:
\begin{enumerate}[leftmargin=*]
    \item Score each point correspondence based on how likely it is to be an inlier.
    \item Choose a sample of 3 points, prioritizing those with higher scores.
    Do not choose a sample that has ever been chosen before.
    \item Check if the sample passes a basic prescreening test. 
    If it passes, continue to the next step.
    Otherwise, go back to the previous step.
    \item Compute the rotation, translation (and scale) between the two sets of 3 points using Horn's method \cite{horn}. 
    \item Find all other points that fit well in the transformation model obtained from the previous step.
    \item Repeat Step 2--5 until a sufficient number of inlier correspondences are found.
    \item Recompute the rotation, translation (and scale) using all the inlier correspondences.
    For this step, we again use Horn's method \cite{horn}.
\end{enumerate}
The novel contribution of this work is an efficient and effective method for Step 1--3.
In the next two sections, we will elaborate on these steps for each unknown-scale and known-scale case separately.
Algorithm \ref{al:unknown} delineates our method for unknown-scale problems.

\begin{algorithm}[t]
\label{al:unknown}
\setstretch{0.88}
\caption{PCR-99 for unknown-scale problems}
\DontPrintSemicolon
\KwInput{Corresponding 3D point sets $\{\mathbf{a}_i\}_{i=1}^{n}$ and $\{\mathbf{b}_i\}_{i=1}^{n}$, Inlier threshold $\delta_\mathrm{in}$.}
\KwOutput{Similarity transformation $(s, \mathbf{R}, \mathbf{t})$.}
\tcc{Compute the log ratio matrix (Sec. \ref{subsec:score}):\hspace{-1em}}

$L \gets \mathbf{0}_{n\times n};$ \label{line:L1}

\For{every possible pair $(i,j)$}
{ 
$L(i,j) \gets \ln\left(\lVert\mathbf{b}_i-\mathbf{b}_j\rVert / \lVert\mathbf{a}_i-\mathbf{a}_j\rVert\right);$ \label{line:L2}
}

\tcc{Score the correspondences (Sec. \ref{subsec:score}):}

$S \gets \mathbf{0}_{1\times n};$ \label{line:S1}

\For{$i=1,2,3, \cdots, n$}
{ 
Obtain $\mathcal{A}$ using \eqref{eq:A1}--\eqref{eq:A2}; \label{line:S1.5}

$\displaystyle S(i) \gets -\hspace{-5pt}\min_{\scaleto{\ln{s}\in\mathcal{A}}{6pt}}\sum_{j=1}^{n} \min\left(\left|L(i,j) - \ln{s}\right|, 0.1\right);$ \label{line:S2}
}

\tcc{Generate and evaluate 3-point samples one by one (Sec. \ref{subsec:ordering} and \ref{subsec:prescreen}):}

By sorting $S$ in a descending order, obtain the ranking numbers $r_i$ for $i=1, 2, \cdots, n$;

$\mathcal{I}_\mathrm{largest}\gets \{\}; \ n_\mathrm{hypothesis}\gets 0;$

\For{$r_\mathrm{sum}=6, 7,  \cdots, 3n-3$ \label{line:so1}}
{
$\left(r_i\right)_\mathrm{L}\gets\max(1, r_\mathrm{sum}-2n+1);$

$\left(r_i\right)_\mathrm{U}\gets\mathrm{floor}\left((r_\mathrm{sum}-3)/3\right);$

\For{$r_i=\left(r_i\right)_\mathrm{L},  \cdots, \left(r_i\right)_\mathrm{U}$}
{

$\left(r_j\right)_\mathrm{L}=\max(r_i+1, r_\mathrm{sum}-r_i-n);$

$\left(r_j\right)_\mathrm{U}=\mathrm{floor}\left((r_\mathrm{sum}-r_i-1)/2\right);$

\For{$r_j=\left(r_j\right)_\mathrm{L},  \cdots, \left(r_j\right)_\mathrm{U}$}
{
$r_k\gets r_\mathrm{sum}-r_i-r_j;$

Find $(i,j,k)$ that corresponds to $(r_i, r_j, r_k)$; \label{line:so2}

\If{$(i,j,k)$ does not pass all of \eqref{eq:condition1}--\eqref{eq:condition2} \label{line:test1}}
{
    Continue; \label{line:test2}
}

Compute $(s, \mathbf{R}, \mathbf{t})$ from $(i,j,k)$ using \cite{horn};\label{line:horn1}

$\mathcal{I}\gets$ All points that fit well in $(s, \mathbf{R}, \mathbf{t})$ under the given inlier threshold $\delta_\mathrm{in}$;

\If{$n(\mathcal{I}) > n(\mathcal{I}_\mathrm{largest})$}
{
$\mathcal{I}_\mathrm{largest}\gets\mathcal{I};$
}

$n_\mathrm{hypothesis}\gets n_\mathrm{hypothesis}+1;$

\If{$n_\mathrm{hypothesis}$ is a multiple of 1000}
{
\If{$n(\mathcal{I}_\mathrm{largest})\geq\max(9, 0.009n)$}
{
Compute $(s, \mathbf{R}, \mathbf{t})$ from $\mathcal{I}_\mathrm{largest}$ using \cite{horn};\label{line:horn2}

\Return{$(s, \mathbf{R}, \mathbf{t})$}
}
}
}
}
}
\end{algorithm}

\section{UNKNOWN-SCALE REGISTRATION}
\label{sec:unknown_scale}
\subsection{Score Function (Step 1 in Section \ref{sec:overview})}
\label{subsec:score}
Consider two arbitrary point correspondences $(\mathbf{a}_i, \mathbf{b}_i)$ and $(\mathbf{a}_j, \mathbf{b}_j)$.
If both of them are inliers, then
\begin{equation}
    \frac{\lVert\mathbf{b}_i-\mathbf{b}_j\rVert}{\lVert\mathbf{a}_i-\mathbf{a}_j\rVert}\approx s,
\end{equation}
where $s$ is the unknown scale.
Taking the natural logarithm on both sides, we get
\begin{equation}
\label{eq:log_ratio1}
    \ln\frac{\lVert\mathbf{b}_i-\mathbf{b}_j\rVert}{\lVert\mathbf{a}_i-\mathbf{a}_j\rVert}-\ln{s} \approx 0.
\end{equation}
Now, we define the \textit{log ratio} between $i$ and $j$ as follows:
\begin{equation}
\label{eq:log_ratio_def}
    L(i, j) = \ln\frac{\lVert\mathbf{b}_i-\mathbf{b}_j\rVert}{\lVert\mathbf{a}_i-\mathbf{a}_j\rVert}.
\end{equation}
Using this definition, we can rewrite \eqref{eq:log_ratio1} into
\begin{equation}
   L(i,j)-\ln{s} \approx 0.
\end{equation}
Point $i$ is likely to be an inlier if the value of $\left|L(i,j)-\ln{s}\right|$ is small with many other points $j$, implying a high degree of \textit{pairwise scale consistency}.
Following this line of reasoning, one may design a score function like this:
\begin{equation}
\label{eq:score1}
    S(i) = -\sum_{j=1}^{n} \min\left(\left|L(i,j) - \ln{s}\right|, \epsilon\right),
\end{equation}
where $\epsilon$ is some threshold.
We truncate $\left|L(i,j) - \ln{s}\right|$ to limit the influence of outliers.
The negative sign is added because we want the score to be high when this truncated cost is low.
The problem with the score function \eqref{eq:score1} is that the scale ${s}$ is unknown.
We solve this problem by modifying \eqref{eq:score1} as follows:
\begin{equation}
\label{eq:score2}
    S(i) = -\min_{\ln{s}}\sum_{j=1}^{n} \min\left(\left|L(i,j) - \ln{s}\right|, \epsilon\right).
\end{equation}
This means that we implicitly find $s$ such that the score function is maximized.
This usually works well because the score is highest when point $i$ is an inlier, and at the same time, $s$ is accurate.
If either one of the conditions is not met, then the score will generally be low.
In our experiment, we find $S(i)$ by trying out a range of values for $\ln{s} \in \mathcal{A}$ where 
\begin{equation}
\label{eq:A1}
    \mathcal{A}=\{p_i, p_i+\Delta_i, p_i+2\Delta_i, p_i+3\Delta_i, \cdots, q_i\},
\end{equation}
\begin{equation}
    p_i=\min_j L(i,j), \ q_i=\max_j L(i,j),
\end{equation}
\begin{equation}
\label{eq:A2}
    \Delta_i = \frac{q_i-p_i}{\max\left(1, \text{Nearest integer of}\left(\displaystyle\frac{q_i-p_i}{0.1}\right)\right)}.
\end{equation}

We score each point correspondence using \eqref{eq:score2}.
To speed up the process, we precompute $L(i,j)$ for every possible pair $(i,j)$ and store the data in a 2D matrix.
With a slight abuse of notation, we use $L$ to denote this matrix and $L(i,j)$ to denote the entry of index $(i,j)$. 
Since $L(i,j)=L(j,i)$ according to \eqref{eq:log_ratio_def}, precomputing the matrix $L$ allows us to avoid redundant operations.
Also, this data can be reused for the prescreening test, which will be discussed in Section \ref{subsec:prescreen}.
In that section, we will also explain why we chose to take the logarithm in \eqref{eq:log_ratio1}.

In Alg. \ref{al:unknown}, the log ratio matrix $L$ is computed in line \ref{line:L1}--\ref{line:L2}  and the correspondences are scored in line \ref{line:S1}--\ref{line:S2}.

\subsection{Sample Ordering (Step 2 in Section \ref{sec:overview})}
\label{subsec:ordering}
Once every point correspondence is scored, we sort them in a descending order and assign ranking numbers ($r_i$) to them:
The one with the highest score is given a ranking number $1$, and the one with the lowest score $n$.
Then, we evaluate the sample with the smallest sum of ranking numbers first, and the one with the largest sum last.
Note that the smallest sum is $1+2+3=6$ and the largest sum is $(n-2)+(n-1)+n=3n-3$.
In the following, we outline the procedure (which corresponds to Step 2 in Section \ref{sec:overview}):
\begin{enumerate}
    \item [2.1)] If Step 2 has been reached for the very first time, set $r_\mathrm{sum}=6$.
    \item [2.2)]Choose a sample of 3 points $(i,j,k)$ such that $r_i+r_j+r_k = r_\mathrm{sum}$.
    Do not choose a sample that has ever been chosen before.
    If no sample can be chosen and $r_\mathrm{sum}<3n-3$, add $1$ to $r_\mathrm{sum}$ and try again.
    We provide the implementation details in the appendix.
    \item [2.3)]Pass the sample on to Step 3 in Section \ref{sec:overview}.
\end{enumerate}
This procedure corresponds to line \ref{line:so1}--\ref{line:so2} of Alg. \ref{al:unknown}.

\subsection{Prescreening Test (Step 3  in Section \ref{sec:overview})}
\label{subsec:prescreen}
Consider a sample of three point correspondences $(\mathbf{a}_i, \mathbf{b}_i)$, $(\mathbf{a}_j, \mathbf{b}_j)$ and $(\mathbf{a}_k, \mathbf{b}_k)$.
The distances between the points within each set are given by:
\begin{gather}
    a_{ij} = \lVert\mathbf{a}_i-\mathbf{a}_j\rVert, \ b_{ij} = \lVert\mathbf{b}_i-\mathbf{b}_j\rVert,\\
    a_{jk} = \lVert\mathbf{a}_j-\mathbf{a}_k\rVert, \ b_{jk} = \lVert\mathbf{b}_j-\mathbf{b}_k\rVert,\\
    a_{ki} = \lVert\mathbf{a}_k-\mathbf{a}_i\rVert, \ b_{ki} = \lVert\mathbf{b}_k-\mathbf{b}_i\rVert.
\end{gather}
If all of the three correspondences are inliers, then the triangle formed by $(\mathbf{a}_i, \mathbf{a}_j, \mathbf{a}_k)$ would be almost similar to that formed by $(\mathbf{b}_i, \mathbf{b}_j, \mathbf{b}_k)$.
This means that
\begin{equation}
\label{eq:similar1}
    \frac{a_{ij}}{b_{ij}} \approx \frac{a_{jk}}{b_{jk}} \approx \frac{a_{ki}}{b_{ki}},
\end{equation}
and equivalently,
\begin{equation}
\label{eq:similar2}
    \frac{a_{ij}}{a_{jk}} \approx \frac{b_{ij}}{b_{jk}}, \ \frac{a_{jk}}{a_{ki}} \approx \frac{b_{jk}}{b_{ki}}, \ \frac{a_{ij}}{a_{ki}} \approx \frac{b_{ij}}{b_{ki}},
\end{equation}
One could turn \eqref{eq:similar1} and \eqref{eq:similar2} into the following necessary conditions for inliers:
\begin{equation}
\label{eq:similar3}
    \left|\frac{a_{ij}}{b_{ij}} - \frac{a_{jk}}{b_{jk}}\right| < \delta, \ \left|\frac{a_{jk}}{b_{jk}} - \frac{a_{ki}}{b_{ki}}\right| < \delta, \ 
    \left|\frac{a_{ij}}{b_{ij}} - \frac{a_{ki}}{b_{ki}}\right| < \delta, 
\end{equation}
\begin{equation}
        \left|\frac{b_{ij}}{a_{ij}} - \frac{b_{jk}}{a_{jk}}\right| < \delta, \ \left|\frac{b_{jk}}{a_{jk}} - \frac{b_{ki}}{a_{ki}}\right| < \delta, \ 
    \left|\frac{b_{ij}}{a_{ij}} - \frac{b_{ki}}{a_{ki}}\right| < \delta, 
\end{equation}
\begin{equation}
    \left|\frac{a_{ij}}{a_{jk}} - \frac{b_{ij}}{b_{jk}}\right| < \delta, \ \left|\frac{a_{jk}}{a_{ki}} - \frac{b_{jk}}{b_{ki}}\right| < \delta, \ 
    \left|\frac{a_{ij}}{a_{ki}} - \frac{b_{ij}}{b_{ki}}\right| < \delta,
\end{equation}
\begin{equation}
\label{eq:similar4}
    \left|\frac{a_{jk}}{a_{ij}} - \frac{b_{jk}}{b_{ij}}\right| < \delta, \ \left|\frac{a_{ki}}{a_{jk}} - \frac{b_{ki}}{b_{jk}}\right| < \delta, \ 
    \left|\frac{a_{ki}}{a_{ij}} - \frac{b_{ki}}{b_{ij}}\right| < \delta, 
\end{equation}
where $\delta$ is some threshold.
Given samples of triplets $(i,j,k)$, checking \eqref{eq:similar3}--\eqref{eq:similar4} can be useful for selecting potential inliers.
However, this can become computationally demanding as the number of samples increases (\textit{e.g.,} due to high outlier ratios).
To mitigate this issue, we take an alternative approach:
We combine \eqref{eq:similar1} and \eqref{eq:similar2} and rewrite them as follows:
\begin{equation}
    \frac{b_{ij}a_{jk}}{a_{ij}b_{jk}} \approx 1, \ \frac{b_{jk}a_{ki}}{a_{jk}b_{ki}} \approx 1, \ \frac{b_{ij}a_{ki}}{a_{ij}b_{ki}} \approx 1.
\end{equation}
Taking the logarithm on both sides, we get
\begin{gather}
    \ln\left(b_{ij}/a_{ij}\right) - \ln\left(b_{jk}/a_{jk}\right)\approx 0,\label{eq:similar5}\\
    \ln\left(b_{jk}/a_{jk}\right) - \ln\left(b_{ki}/a_{ki}\right)\approx 0,\\
    \ln\left(b_{ij}/a_{ij}\right) - \ln\left(b_{ki}/a_{ki}\right)\approx 0.\label{eq:similar6}
\end{gather}
We turn \eqref{eq:similar5}--\eqref{eq:similar6} into the following necessary conditions for inliers:
\begin{gather}
    \left|L(i,j) - L(j,k)\right| < \epsilon,\label{eq:condition1}\\
    \left|L(i,k) - L(k,i)\right| < \epsilon,\\
    \left|L(i,j) - L(k,i)\right| < \epsilon,\label{eq:condition2}
\end{gather}
where $\epsilon$ is the same threshold\footnote{In our experiments, we set $\epsilon=0.1$.} used for \eqref{eq:score1} and \eqref{eq:score2} and $L(\cdot, \cdot)$ is the log ratio defined in \eqref{eq:log_ratio_def}.
Essentially, these conditions check the \textit{triplet scale consistency} of a sample $(i, j, k)$.
Since we have already computed and stored the log ratios in matrix $L$ (as discussed in Section \ref{subsec:score}), checking \eqref{eq:condition1}--\eqref{eq:condition2} is much faster than checking \eqref{eq:similar3}--\eqref{eq:similar4}\footnote{The proposed conditions \eqref{eq:condition1}--\eqref{eq:condition2} are not necessarily equivalent to \eqref{eq:similar3}--\eqref{eq:similar4}.
In practice, one may use either set of conditions by specifying appropriate thresholds $\delta$ and $\epsilon$.
We simply chose to use \eqref{eq:condition1}--\eqref{eq:condition2} for efficiency reasons.}.
This is why we use the log ratio for the prescreening test, as well as the score function \eqref{eq:score2}. 
The prescreening test is done in line \ref{line:test1}--\ref{line:test2} of Alg. \ref{al:unknown}.

{\renewcommand{\arraystretch}{1.3}%
\begin{table}[t]
\small
\begin{center}
\begin{tabular}{cclrr}
\hline
 & & \multicolumn{1}{c}{} & \multicolumn{1}{c}{$\mathbf{R}$ error $>5^\circ$} & \multicolumn{1}{c}{$\mathbf{R}$ error $>10^\circ$} \\
 \hline
 \multicolumn{1}{c}{\parbox[t]{1mm}{\multirow{4}{*}{\rotatebox[origin=c]{90}{Unknown}}}} & \multicolumn{1}{c}{\hspace{-3mm}\parbox[t]{1mm}{\multirow{4}{*}{\rotatebox[origin=c]{90}{scale}}}} & RANSIC \cite{ransic} & 79 out of 5500 & 79 out of 5500 \\
 \multicolumn{1}{c}{}&\multicolumn{1}{c}{} & ICOS \cite{icos} & 12 out of 5500 & 11 out of 5500\\
 \multicolumn{1}{c}{}&\multicolumn{1}{c}{} & PCR-99 (no S/O) & 1 out of 5500 & \textbf{0 out of 5500}\\
 \multicolumn{1}{c}{}&\multicolumn{1}{c}{} & PCR-99 & \textbf{0 out of 5500} & \textbf{0 out of 5500}\\
 \hline
 \multicolumn{1}{c}{\parbox[t]{1mm}{\multirow{8}{*}{\rotatebox[origin=c]{90}{Known}}}} & \multicolumn{1}{c}{\hspace{-3mm}\parbox[t]{1mm}{\multirow{8}{*}{\rotatebox[origin=c]{90}{scale}}}} & RANSIC \cite{ransic} &81 out of 5500&78 out of 5500\\
 \multicolumn{1}{c}{}&\multicolumn{1}{c}{} & ICOS \cite{icos} & 24 out of 5500& 22 out of 5500\\
 \multicolumn{1}{c}{}&\multicolumn{1}{c}{} & DANIEL \cite{daniel} & 5 out of 5500 & 3 out of  5500 \\
 \multicolumn{1}{c}{}&\multicolumn{1}{c}{} & TriVoC \cite{trivoc} & \textbf{2 out of  5500}& \textbf{0 out of  5500}\\
 \multicolumn{1}{c}{}&\multicolumn{1}{c}{} & VODRAC \cite{vodrac} &9 out of  5500&2 out of  5500\\
 \multicolumn{1}{c}{}&\multicolumn{1}{c}{} & VOCRA \cite{vocra} &16 out of  5500& 16 out of  5500\\
 \multicolumn{1}{c}{}&\multicolumn{1}{c}{} & PCR-99 (no S/O) &4 out of  5500& \textbf{0 out of  5500}\\
 \multicolumn{1}{c}{}&\multicolumn{1}{c}{} & PCR-99 &4 out of  5500& \textbf{0 out of  5500}\\
 \hline
 \end{tabular}
\end{center}
\caption{The number of times large errors are produced.
PCR-99 always achieves the best or the second best results.}
\label{tab:large_errors}
\end{table}
}
{\renewcommand{\arraystretch}{1.3}%
\begin{table}[t]
\small
\begin{center}
\begin{tabular}{cclc}
\hline
 & & \multicolumn{1}{c}{} & \multicolumn{1}{c}{Speedup}  \\
 \hline
 \multicolumn{1}{c}{\parbox[t]{1mm}{\multirow{3}{*}{\rotatebox[origin=c]{90}{Unknown}}}} & \multicolumn{1}{c}{\hspace{-3mm}\parbox[t]{1mm}{\multirow{3}{*}{\rotatebox[origin=c]{90}{scale}}}} & RANSIC \cite{ransic} & 16  \\
 \multicolumn{1}{c}{}&\multicolumn{1}{c}{} & ICOS \cite{icos} & 7.0 \\
 \multicolumn{1}{c}{}&\multicolumn{1}{c}{} & PCR-99 (no S/O) & 1.6 \\
 \hline
 \multicolumn{1}{c}{\parbox[t]{1mm}{\multirow{7}{*}{\rotatebox[origin=c]{90}{Known}}}} & \multicolumn{1}{c}{\hspace{-3mm}\parbox[t]{1mm}{\multirow{7}{*}{\rotatebox[origin=c]{90}{scale}}}} & RANSIC \cite{ransic} & 162\\
 \multicolumn{1}{c}{}&\multicolumn{1}{c}{} & ICOS \cite{icos} & 33\\
 \multicolumn{1}{c}{}&\multicolumn{1}{c}{} & DANIEL \cite{daniel} & 8.0 \\
 \multicolumn{1}{c}{}&\multicolumn{1}{c}{} & TriVoC \cite{trivoc} & 1.4\\
 \multicolumn{1}{c}{}&\multicolumn{1}{c}{} & VODRAC \cite{vodrac} & 3.0\\
 \multicolumn{1}{c}{}&\multicolumn{1}{c}{} & VOCRA \cite{vocra} & 1.2\\
 \multicolumn{1}{c}{}&\multicolumn{1}{c}{} & PCR-99 (no S/O) & 9.3\\
 \hline
 \end{tabular}
\end{center}
\caption{Median speedup of PCR-99 compared 
to the other methods at 99\% outlier ratio.}
\label{tab:timings}
\end{table}
}

\section{KNOWN-SCALE REGISTRATION}
\label{sec:known_scale}
When the relative scale $s$ between the two 3D point sets is known, we simply modify our method as follows:
\begin{enumerate}
    \item We use \eqref{eq:score1} instead of \eqref{eq:score2} for the score function, replacing line \ref{line:S1.5}--\ref{line:S2} of Alg. \ref{al:unknown}.
    \item We replace the prescreening conditions \eqref{eq:condition1}--\eqref{eq:condition2} in line \ref{line:test1} of Alg. \ref{al:unknown} with 
    \begin{gather}
    \left|L(i,j) - \ln{s}\right| < \epsilon,\\
    \left|L(j,k) - \ln{s}\right| < \epsilon,\\
    \left|L(k,i) - \ln{s}\right| < \epsilon.
    \end{gather}
    \item In line \ref{line:horn1} and \ref{line:horn2} of Alg. \ref{al:unknown}, we do not compute $s$. 

\end{enumerate}

\section{RESULTS}
\label{sec:results}

We compare our method, PCR-99, against six open-source state-of-the-art methods: RANSIC \cite{ransic}, ICOS \cite{icos}, DANIEL \cite{daniel}, TriVoC \cite{trivoc}, VODRAC \cite{vodrac} and VOCRA \cite{vocra}.
We chose these methods for benchmarking, as they are claimed to be fast and robust up to 99\% outliers.
RANSAC \cite{ransac}, FGR \cite{fgr}, GORE \cite{bustos_2018_tpami}, GNC-TLS \cite{gnc} and TEASER \cite{teaser} are not included in the evaluation, as they are shown to be outperformed by the aforementioned methods (when parallelization is disabled for the fair runtime evaluation) \cite{ransic, icos, daniel, trivoc, vodrac, vocra}.
We also compare our method with an alternative version that adopts random sampling (as in RANSAC \cite{ransac}) instead of the method described in Section \ref{subsec:ordering}. 
At 99\% outlier ratio, RANSIC and ICOS sometimes take hours to find a good solution.
To prevent excessive computation times, we limit the iterations to 100 seconds for all methods.
For known-scale problems, we evaluate all of the aforementioned methods.
For unknown-scale problems, we only evaluate PCR-99 (with or without sample ordering), RANSIC and ICOS, as the other four methods are not applicable.
All experiments are conducted in MATLAB on a laptop with i7-7700HQ (2.8 GHz) and 16 GB RAM.

\begin{figure*}[t]
 \centering
 \includegraphics[width=\textwidth]{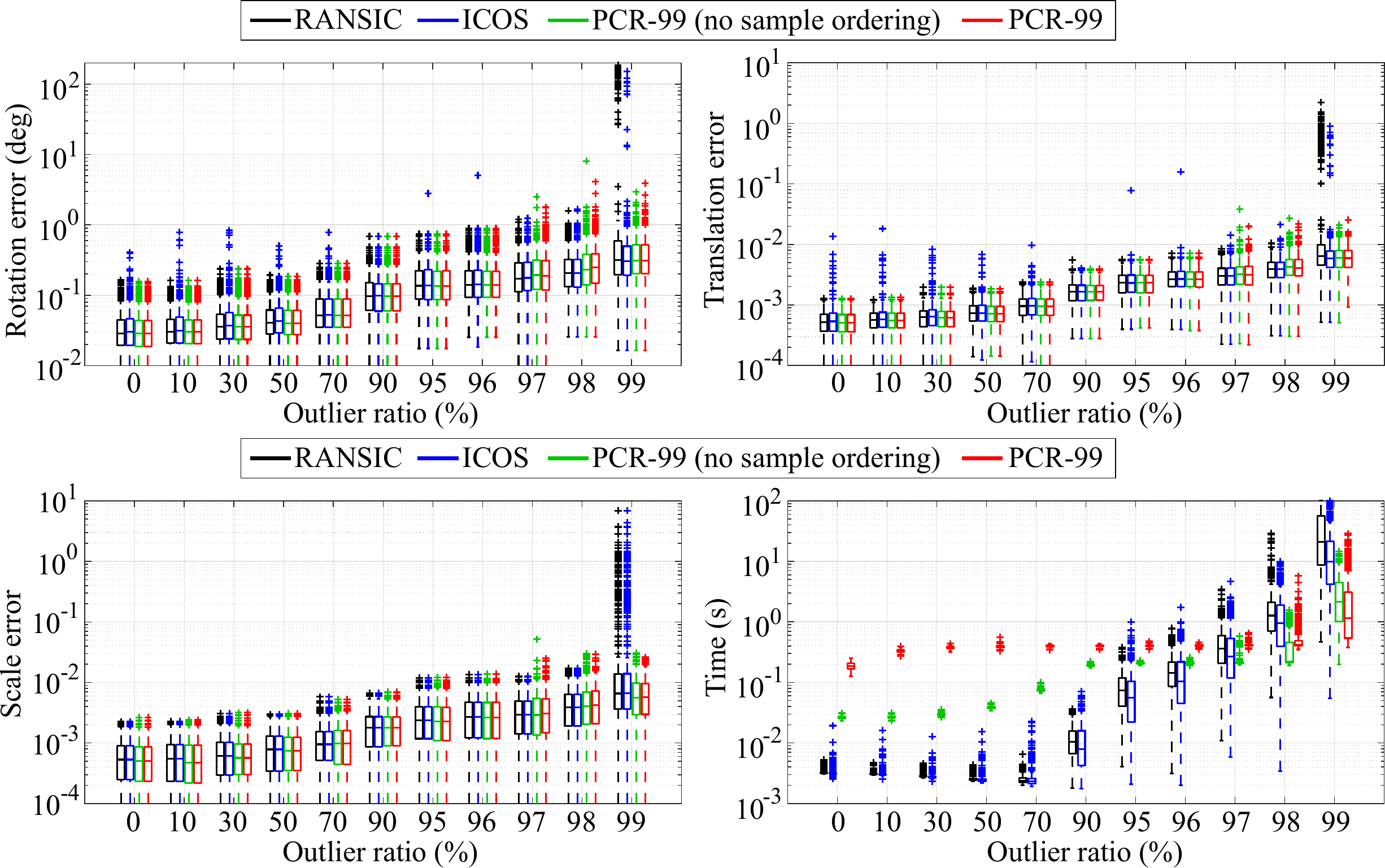}
\caption{Evaluation results for \textbf{unknown-scale} problems (500 Monte Carlo runs): RANSIC and ICOS are not robust at 99\% outlier ratio when the computation time is limited to 100 seconds.
In contrast, PCR-99 (with or without sample ordering) is robust up to 99\% outlier ratio.
The sample ordering has a very small impact on the accuracy and robustness of PCR-99.
Comparing the median times of the two versions of PCR-99, the random sampling approach is faster up to 98\% outlier ratio, but slower at 99\% outlier ratio.}
\label{fig:results_unscaled}
\end{figure*}

\subsection{Unknown-Scale Registration}
We use the Bunny dataset from the Stanford 3D scanning repository \cite{bunny}.
The dataset is processed as follows:
First, the ground-truth point cloud is obtained by downsampling the data to 1000 points and scaling it to fit inside a unit cube.
To obtain the query point cloud, we first transform  the ground truth using a random similarity transformation with a scale $s$ where $1<s<5$.
We then add Gaussian noise of $\sigma=0.01$ and replace 0--99\% of the points with random points inside a 3D sphere of diameter $\sqrt{3}s$ (see Fig. \ref{fig:bunny}).

Fig. \ref{fig:results_unscaled} presents the results. 
At 99\% outlier ratio, PCR-99 (with or without sample ordering) is the only method that achieves robust registration in 100 seconds at all times (see also Tab.\ref{tab:large_errors}).
We provide the relative speed of PCR-99 with respect to the other methods in Tab. \ref{tab:timings}.

\subsection{Known-Scale Registration}
We follow the same procedure as in the previous section to simulate the data for known-scale problems.
The only difference here is that we set $s=1$ at all times.

\begin{figure*}[t]
 \centering
 \includegraphics[width=\textwidth]{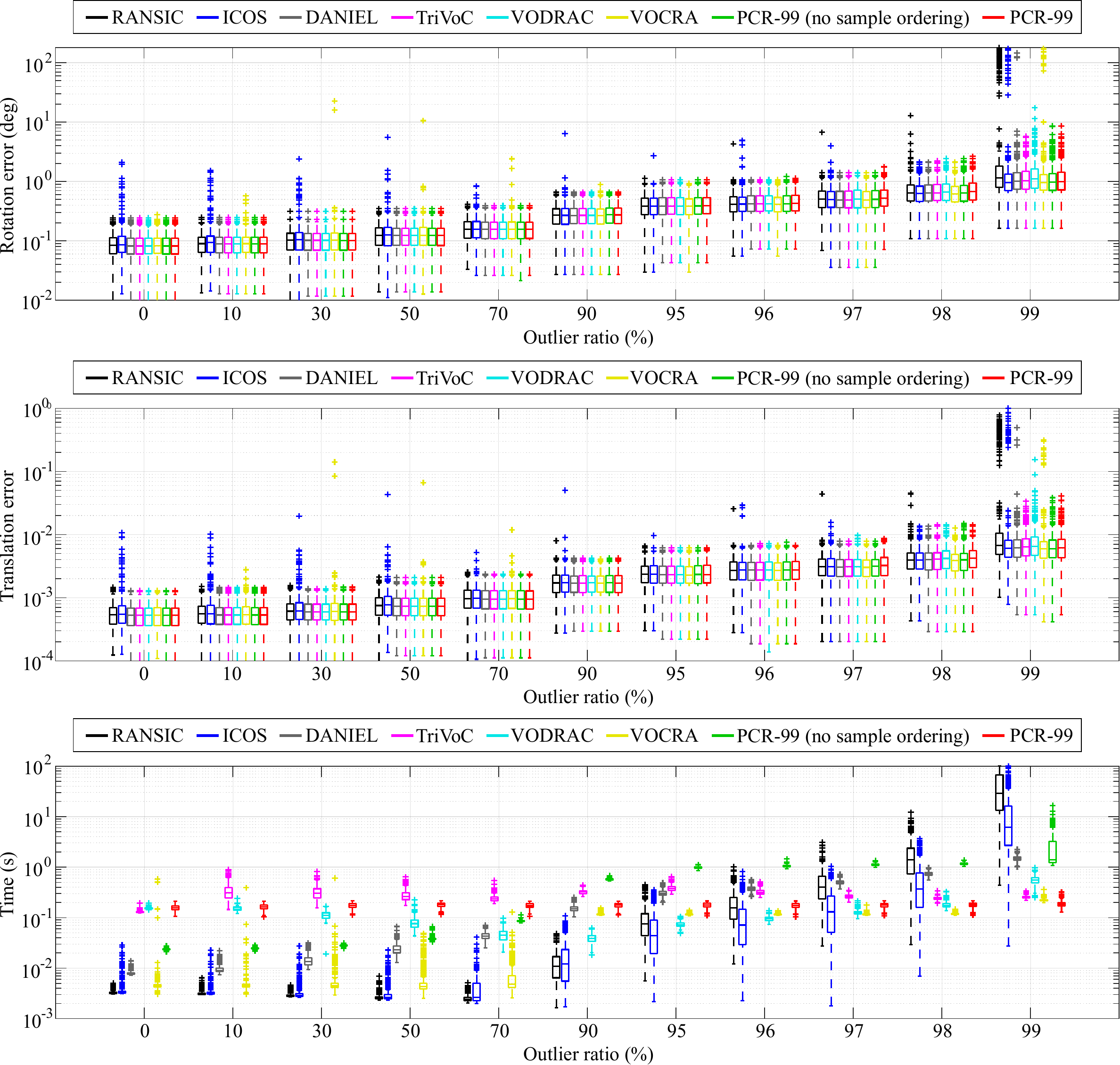}
\caption{Evaluation results for \textbf{known-scale} problems (500 Monte Carlo runs): TriVoC, VODRAC, and PCR-99 (with or without sample ordering) are more robust than the other methods, producing fewer large rotation errors (\textit{e.g.}, larger than $10^\circ$).
They exhibit similar levels of robustness across all outlier ratios.
For PCR-99, the sample ordering boosts the speed significantly at high outlier ratios, without causing a noticeable change in the accuracy or robustness.
Among all methods, PCR-99 is the fastest one at 99\% outlier ratio.}
\label{fig:results_scaled}
\end{figure*}

Fig. \ref{fig:results_scaled} presents the results.
At 99\% outlier ratio, TriVoC, VODRAC and PCR-99 (with or without sample ordering) are the three most robust methods (see also Tab. \ref{tab:large_errors}).
In Tab. \ref{tab:timings}, we also show that PCR-99 is the fastest method at this extreme outlier ratio.

\section{CONCLUSIONS}
\label{sec:conclusion}
In this work, we proposed PCR-99, a novel method for robust point cloud registration that is capable of handling extremely large outlier ratios in an efficient manner.
We first score each point correspondence based on how likely it is to be an inlier.
Specifically, we adopt a novel score function that evaluates the pairwise scale consistency.
After ranking the correspondences, we evaluate a sequence of 3-point samples, starting from the one with the smallest total ranking number and gradually moving towards a larger and larger number. 
This approach allows us to prioritize samples that are more likely to consist of inliers only.
Finally, we perform a prescreening test to check the triplet scale consistency and reject bad samples as early as possible.
Once we find a good sample that leads to a sufficient number of inliers, we recompute the transformation using the inlier set.

For unknown-scale problems, we demonstrated that PCR-99 is as robust as the state of the art up to 98\% outlier ratio and significantly more robust and faster at 99\% outlier ratio.
For known-scale problems, it is as robust as the state of the art, but faster at 99\% outlier ratio.





\section*{APPENDIX}

Here, we elaborate on the implementation details of Step 2.2 in Section \ref{subsec:ordering}.
We want to find three ranking numbers $r_i$, $r_j$ and $r_k$ (where $r_i<r_j<r_k$) such that they add up to a given value, $r_\mathrm{sum}$. 
They have the following constraints:
\begin{equation}
\label{eq:app1}
    r_i+r_j+r_k = r_\mathrm{sum}, 
\end{equation}
\begin{equation}
\label{eq:app2}
    1 \leq r_i, 
\end{equation}
\begin{equation}
\label{eq:app3}
    r_i+1 \leq r_j,
\end{equation}
\begin{equation}
\label{eq:app4}
    r_j+1 \leq r_k, 
\end{equation}
\begin{equation}
\label{eq:app5}
     r_i+2 \leq r_k,
\end{equation}
\begin{equation}
 \label{eq:app6}
    r_k \leq n,
\end{equation}
\begin{equation}
 \label{eq:app7}
    r_j \leq n-1,
\end{equation}
\begin{equation}
\label{eq:app8}
    r_i \leq n-2. 
\end{equation}
In the remainder of this section, we will manipulate \eqref{eq:app1}--\eqref{eq:app8} to obtain lower and upper bounds of $r_i$, $r_j$ and $r_k$ we can use in practice.

Adding $r_i$ to both sides of \eqref{eq:app3} and adding \eqref{eq:app5}, we get
\begin{equation}
    3r_i+3 \leq r_i+r_j+r_k\stackrel{\eqref{eq:app1}}{=}r_\mathrm{sum}.
\end{equation}
\begin{equation}
\label{eq:app9}
    \Rightarrow r_i\leq(r_\mathrm{sum}-3)/3.
\end{equation}
Adding $r_j$ to both sides of \eqref{eq:app4}, we get
\begin{gather}
    2r_j+1 \leq r_j+r_k\stackrel{\eqref{eq:app1}}{=}r_\mathrm{sum}-r_i.\\
    \Rightarrow r_j\leq(r_\mathrm{sum}-r_i-1)/2. \label{eq:app10}
\end{gather}
Adding \eqref{eq:app6} and \eqref{eq:app7} together, we get
\begin{gather}
    r_j+r_k \leq 2n-1 \ 
    \stackrel{\eqref{eq:app1}}{\Rightarrow} \ r_\mathrm{sum}-r_i \leq 2n-1.\\
    \Rightarrow r_\mathrm{sum}-2n+1 \leq r_i. \label{eq:app11}\\
    \Rightarrow (r_\mathrm{sum}-r_i-1)/2 \leq n-1.\label{eq:app12}
\end{gather}
Rewriting \eqref{eq:app6} using \eqref{eq:app1}, we get
\begin{gather}
    r_\mathrm{sum}-r_i-r_j \leq n \ 
    \Rightarrow \ r_\mathrm{sum}-r_i-n \leq r_j. \label{eq:app13}
\end{gather}
Adding \eqref{eq:app6}--\eqref{eq:app8} altogether, we get 
\begin{equation}
\label{eq:app14}
    r_\mathrm{sum}\leq 3n-3 \ \Rightarrow \ (r_\mathrm{sum}-3)/3 \leq n-2.
\end{equation}

Now we are ready to derive the lower and upper bounds for $r_i$ and $r_j$ which can be used in practice:
Combining \eqref{eq:app2}, \eqref{eq:app8}, \eqref{eq:app9} and \eqref{eq:app11}, we get
\begin{equation}
    \label{eq:app15}
    \max(1, r_\mathrm{sum}-2n+1) \leq r_i \leq \min\left(n-2, (r_\mathrm{sum}-3)/3\right).
\end{equation}
Combining \eqref{eq:app14} and \eqref{eq:app15}, we get
\begin{equation}
    \label{eq:r_i}
    \max(1, r_\mathrm{sum}-2n+1) \leq r_i \leq (r_\mathrm{sum}-3)/3.
\end{equation}
Combining \eqref{eq:app3}, \eqref{eq:app7}, \eqref{eq:app10} and \eqref{eq:app13}, we get
\begin{equation}
\label{eq:app16}
    \max(r_i+1, r_\mathrm{sum}-r_i-n) \leq r_j \leq \min\hspace{-2pt}\left(\hspace{-2pt}n-1, \frac{r_\mathrm{sum}-r_i-1}{2}\right)
\end{equation}
Combining \eqref{eq:app12} and \eqref{eq:app16}, we get
\begin{equation}
\label{eq:r_j}
        \max(r_i+1, r_\mathrm{sum}-r_i-n) \leq r_j \leq (r_\mathrm{sum}-r_i-1)/2.
\end{equation}

The inequality \eqref{eq:r_i} is useful because it gives us the lower and upper bound of $r_i$ independently of $r_j$ and $r_k$.
The inequality \eqref{eq:r_j} is also useful because it allows us to compute the lower and upper bound of $r_j$ once $r_i$ is determined.
Once both $r_i$ and $r_j$ are determined, then $r_k$ is automatically determined using \eqref{eq:app1}.
Given a fixed value for $r_\mathrm{sum}$, we produce a sequence of triplets $(r_i,r_j,r_k)$ as follows:
\begin{enumerate}
    \item In the outer loop, let $r_i$ iterate from the lower bound $\left(r_i\right)_\mathrm{L}$ to the upper bound $\left(r_i\right)_\mathrm{U}$, where
    \begin{gather}
    \left(r_i\right)_\mathrm{L}=\max(1, r_\mathrm{sum}-2n+1),\\
\left(r_i\right)_\mathrm{U}=\mathrm{floor}\left((r_\mathrm{sum}-3)/3\right).
    \end{gather}
    \item In the inner loop, let $r_j$ iterate from
    the lower bound $\left(r_j\right)_\mathrm{L}$ to the upper bound $\left(r_j\right)_\mathrm{U}$, where
    \begin{gather}
    \left(r_j\right)_\mathrm{L}=\max(r_i+1, r_\mathrm{sum}-r_i-n),\\
    \left(r_j\right)_\mathrm{U}=\mathrm{floor}\left((r_\mathrm{sum}-r_i-1)/2\right),
    \end{gather}
    Once $r_i$ and $r_j$ are determined, we can determine $r_k$:
    \begin{equation}
        r_k = r_\mathrm{sum}-r_i-r_j.
    \end{equation}
\end{enumerate}




\end{document}